\documentclass[10pt,twocolumn,letterpaper]{article}
\usepackage[pagenumbers]{cvpr}
\usepackage{tabularx,array,algorithm,algpseudocode,cuted,needspace}
\definecolor{cvprblue}{rgb}{0.21,0.49,0.74}
\usepackage[breaklinks,colorlinks,allcolors=cvprblue]{hyperref}
\title{Multi-Teacher On-Policy Distillation for Flow-Based Vision-Language-Action Models}
\author{\textbf{Lichao Ni}$^{1,2}$\quad \textbf{Yuhan Liu}$^{1}$\quad \textbf{Pengfei Li}$^{1}$\quad \textbf{Guanghuan Fang}$^{1,\dagger}$\\[4pt]$^{1}$Huawei Technologies Co., Ltd.\quad $^{2}$Tsinghua University}
\hypersetup{pdftitle={Multi-Teacher On-Policy Distillation for Flow-Based Vision-Language-Action Models},pdfauthor={Lichao Ni, Yuhan Liu, Pengfei Li, Guanghuan Fang}}
\newcommand{\E}{\mathbb{E}}

\newcommand{\Normal}{\mathcal{N}}
\newcommand{\Loss}{\mathcal{L}}
\newcommand{\sg}{\operatorname{sg}}
\newcommand{\MSE}{\operatorname{MSE}}

\newcommand{\pizero}{\ensuremath{\pi_{0}}}
\newcommand{\pihalf}{\ensuremath{\pi_{0.5}}}

\begin{document}
\maketitle
\begingroup\renewcommand{\thefootnote}{\fnsymbol{footnote}}\footnotetext[2]{Corresponding author.}\endgroup

\begin{abstract}
We study how to distill multiple vision-language-action (VLA) specialists into a single policy for robot manipulation. In a flow-based VLA, each action chunk is generated along a trajectory through intermediate flow states, and the actions executed in the environment determine which observations the student encounters next. We introduce \emph{Velocity Loss MOPD}, a multi-teacher on-policy distillation method that queries the frozen domain specialist at each observation the student visits and supplies a target velocity at every nonterminal solver step along the path that produced the student's actions. After 80 updates from a shared supervised initialization, Velocity Loss MOPD outperforms both Action Loss OPD and conditional flow matching on teacher-generated actions, with gains of 6.7--9.5 percentage points on LIBERO-40, 3.4--4.2 on Meta-World MT50, and 4.8--18.3 on CALVIN five-instruction completion. Moreover, the advantage of student sampling is especially clear on long-horizon tasks. In our $2\times2$ ablation, using the student for both environment rollouts and flow-state generation achieves higher success on LIBERO-10 and five-instruction CALVIN sequences than using the teacher for either or both.
\end{abstract}

\begin{figure*}[!t]
\centering
\IfFileExists{cvpr2027/figures/framework.pdf}{%
  \includegraphics[width=\textwidth,height=58mm,keepaspectratio]{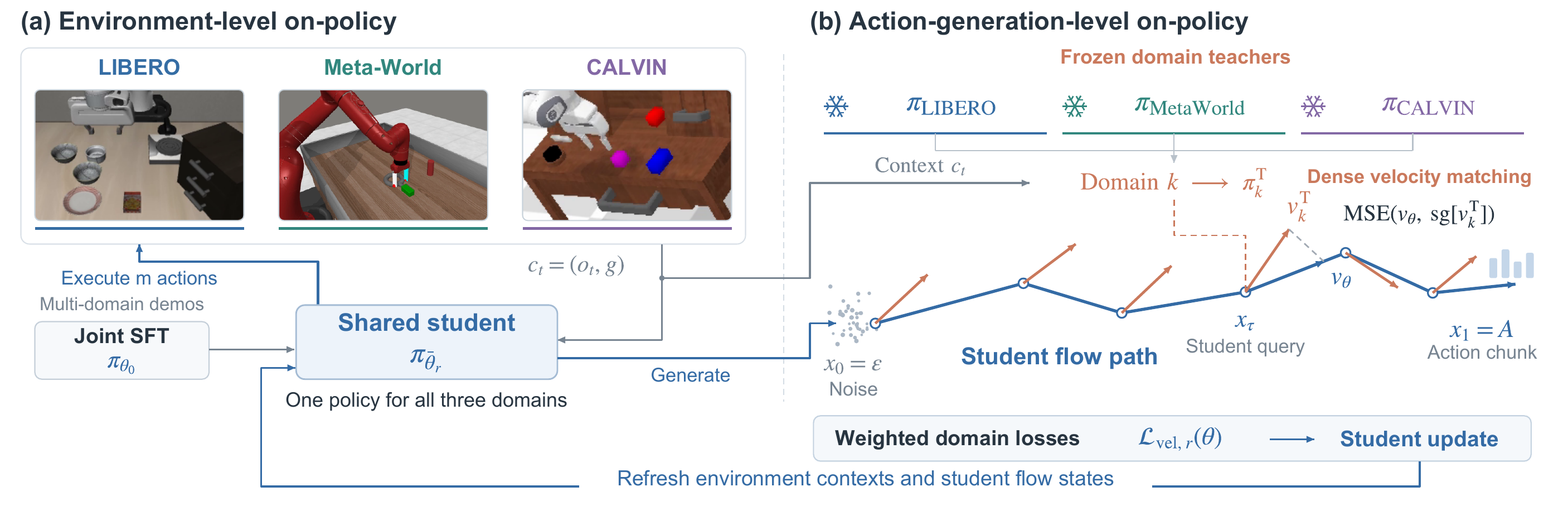}
  \caption{\textbf{Two-level multi-teacher on-policy distillation.} (a) A shared student initialized by joint SFT interacts with three manipulation domains to collect closed-loop decision contexts $c_t=(o_t,g)$, which are routed to both the student and the domain-matched specialist teacher. (b) Along the student's flow path from noise to an action chunk, the teacher provides dense velocity targets (orange vectors) at every nonterminal solver state. Query states and teacher targets are detached via stop-gradient ($\sg$) during optimization, and weighted domain losses update the shared student before the next round of environment and flow-state collection.}
}{%
  \begingroup
  \setlength{\fboxsep}{0pt}
  \setlength{\fboxrule}{0.4pt}
  \fcolorbox{black!25}{white}{\rule{0pt}{46mm}\rule{\dimexpr\textwidth-2\fboxrule\relax}{0pt}}
  \endgroup
  \caption{Framework overview. Diagram to be inserted.}
}
\label{fig:framework}
\end{figure*}

\section{Introduction}
\label{sec:intro}

Pretrained vision-language-action (VLA) models such as OpenVLA, \pizero, and \pihalf\ provide strong visuomotor priors for manipulation \citep{kim2024openvla,black2024pi0,physicalintelligence2025pi05}, yet achieving high task competence across diverse environments typically requires domain-specific post-training via supervised fine-tuning or online reinforcement learning \citep{chen2025pirl}. Deploying the specialists separately requires maintaining a dedicated model for each domain. Multi-teacher policy consolidation instead merges these specialists into a single shared VLA policy operating across distinct observation and action interfaces.

Off-policy distillation from expert demonstrations \citep{xu2025rldg} suffers from compounding covariate shift once the student deviates from expert trajectories: a missed grasp alters both the next visual observation and the corrective action required to recover. On-policy distillation (OPD) addresses environment-level shift by querying teachers on student-visited states \citep{ross2011dagger,zhong2026vlaopd}. In flow-based VLAs, however, generating an action chunk is itself an iterative process that integrates a velocity field from Gaussian noise to control actions, inducing a second, internal distribution of intermediate flow states where teacher guidance must match student behavior.

Effective distillation therefore depends on both \emph{what} target the teacher supplies and \emph{where} supervision is queried. At a student-visited observation, a specialist can supervise only the terminal action chunk via endpoint regression, construct synthetic linear interpolants between noise and teacher actions via conditional flow matching (CFM), or predict local velocities directly along the student's integration path. Prior work studies teacher supervision along student-generated paths in continuous generators and embodied policies \citep{fang2026flowopd,oh2026fastopd,liu2026wamopd}. Here we study where to query domain specialists as a shared policy learns to act across environments.

We propose \emph{Velocity Loss MOPD}, a two-level framework in which frozen domain specialists supervise one shared student along its own environment and flow trajectories (Figure~\ref{fig:framework}). Whenever the student generates an action chunk at a visited observation, its domain teacher labels every nonterminal state on that flow path; holding those states and teacher targets fixed gives a local velocity regression at each solver node before weighted domain losses update the shared student for the next round of closed-loop collection.

From the same SFT checkpoint, Velocity Loss MOPD achieves higher success rates at update 80 than both endpoint action matching and Teacher-action CFM on all three benchmarks. Our $2\times2$ comparison shows that the effect of switching to student environment rollouts depends on which policy generates the flow states used for supervision.

\Needspace{5\baselineskip}
Our contributions are threefold:
\begin{itemize}[leftmargin=*,itemsep=2pt,topsep=3pt]
\item We present a two-level multi-teacher OPD framework (\emph{Velocity Loss MOPD}) that couples closed-loop student environment collection with dense teacher-velocity supervision on the flow paths the student uses to generate its action chunks.
\item We consolidate three manipulation specialists into one shared \pihalf\ policy that is competitive with independent single-teacher distillation under matched per-domain trajectory budgets. Distillation raises success rates over joint SFT by 22.55 percentage points on LIBERO-40, 22.6 on MT50, and 33.4 on CALVIN Len-5.
\item We compare endpoint action matching, Teacher-action CFM, and velocity matching along student flow paths, and run a $2\times2$ ablation (comparing S/S against T/T, T/S, and S/T, where S denotes the student and T the teacher) that holds the velocity loss fixed while varying which policy collects the environment rollouts and which generates the flow states.
\end{itemize}

\section{Related Work}
\label{sec:related}

\begin{algorithm*}[!t]
\caption{Multi-teacher on-policy distillation with Velocity Loss}
\label{alg:training}
\small
\algrenewcommand{\algorithmicrequire}{\textbf{Input:}}
\algrenewcommand{\algorithmicensure}{\textbf{Output:}}
\begin{algorithmic}[1]
\Require SFT initialization $\theta_0$, frozen teachers $\{\pi_k^{\mathrm{T}}\}$, episodes per domain $n$, rounds $R$.
\Ensure One shared student $\pi_\theta$.
\State Initialize $\theta\gets\theta_0$.
\For{$r=0,\ldots,R-1$}
  \State Freeze snapshot $\bar\theta\gets\sg[\theta]$ and clear accumulated gradients.
  \For{each domain $k$}
    \State \textit{Student sampling:} collect $n$ episodes using $\pi_{\bar\theta}$.
    \State Retain supervision contexts and student-generated intermediate flow states.
    \State \textit{Teacher labeling:} query $\pi_k^{\mathrm{T}}$ at these states and freeze its velocity targets.
    \State Accumulate the weighted Velocity Loss gradients using Equation~\eqref{eq:opd}.
  \EndFor
  \State \textit{Student update:} clip gradients and take one AdamW step after all domains are processed.
\EndFor
\end{algorithmic}
\end{algorithm*}

\paragraph{On-policy distillation.}
In sequential control, a learner's actions govern the future states it visits, causing offline imitation to suffer from compounding distribution shift. DAgger resolves this mismatch by querying an expert on learner-visited states \citep{ross2011dagger}, and policy-distillation analyses formalize how sampling distributions and divergence choices shape convergence \citep{czarnecki2019distilling}. This principle underpins post-training for autoregressive language models, including matching teacher token distributions on student continuations in GKD \citep{agarwal2024gkd}, optimizing sequence-level reverse KL via policy gradients in MiniLLM \citep{gu2024minillm}, and applying token-level reverse KL on self-generated reasoning traces \citep{lu2025onpolicydistillation}.

\paragraph{Learning from multiple specialists.}
Consolidating multiple specialists into one policy dates back to multi-task policy distillation \citep{rusu2016policy} and has been scaled to language models in MOPD \citep{ma2026mopd}. To mitigate cross-domain interference, recent studies balance progress via loss allocation and dynamic scheduling \citep{gao2026openmopd,sun2026d3mopd}, domain-normalized feedback and update projection \citep{li2026dnmopd,zhu2026upmopd}, or trajectory filtering in offline multi-teacher distillation \citep{xie2026rethinking}. In robotics, RLDG distills trajectories from task-specific RL specialists \citep{xu2025rldg}, whereas PLD collects recovery data by letting residual specialists intervene during base-policy execution \citep{xiao2026pld}. Related approaches decouple reasoning and action teachers \citep{fang2025dualvla}, query navigation experts online \citep{xu2025mmnav}, merge specialists in parameter space \citep{fu2025mergevla}, or weight specialist actions with phase-specific critics on learner states in REPAIR \citep{jiang2026repair}. Rather than pooling offline rollouts or reweighting candidate actions, we consolidate manipulation specialists---including policies trained with online RL \citep{chen2025pirl}---by matching their velocity fields along student-induced trajectories.

\paragraph{Distillation along the action-generation path.}
Diffusion policies \citep{chi2023diffusionpolicy} and conditional flow matching \citep{lipman2023flowmatching} map noise to action chunks through iterative integration, exposing intermediate states where a teacher can guide generation. In image generation, DiffusionOPD aligns teacher transitions on student sampling paths \citep{li2026diffusionopd}, Flow-OPD adds manifold anchoring \citep{fang2026flowopd}, and FlowMap-OPD separates rollout distributions from transition kernels \citep{li2026flowmapopd}. For robot policies, \citet{deshmukh2026multitaskcfm} distill single-task CFM experts into one multi-task policy by matching expert velocities on linear interpolants between noise and demonstration actions, anchored to ground-truth CFM on fixed demonstration states. To speed up inference, other work reduces transformer depth \citep{jeon2026shallowpi} or compresses multi-step flow models into one-step generators \citep{luan2026snapflow,jun2026driftopd}. Closer to internal-state alignment, FastOPD \citep{oh2026fastopd} queries teacher velocities at a single student-generated flow state and adds a self-consistency loss for few-step deployment, though it still conditions generation only on fixed dataset observations rather than closed-loop rollouts.

\paragraph{On-policy supervision in closed-loop control.}
In closed-loop control, action-generation errors alter executed actions and shift subsequent observations. VLA-OPD applies token-level reverse KL on student rollouts \citep{zhong2026vlaopd}, while related methods focus on quantization-vulnerable states \citep{ruan2026pivotq} or use teacher uncertainty to guide supervision \citep{ren2026uaopd}. TILDE pools rollouts from a fixed behavior-cloned student with teacher recoveries and matches \pihalf\ velocity predictions under shared noise \citep{lu2026tilde}. Two concurrent WAM-OPD studies also separate environment history from the action-generation path \citep{liu2026wamopd,yang2026wamopd}. \citet{yang2026wamopd} keep collected student histories fixed, build velocity targets from teacher actions and initial noise, and supervise both predicted video and actions. \citet{liu2026wamopd} replay stored histories from a fixed pool, query multiple task teachers along the low-noise portion of fresh student denoising paths, and also evaluate an online collection variant. Building on these works, our shared student collects fresh closed-loop rollouts across all three domains after every update, and each domain specialist provides velocity targets at every nonterminal solver state along the path that generated the executed chunk.

\setcounter{dbltopnumber}{1}
\begin{table*}[t]
\centering
\caption{Reference policies and distillation results at round 80. The first two rows use three domain-specific models each. All other rows use one shared student. Velocity loss (S/S, ours) is Velocity Loss MOPD, with student rollouts and flow states. Bold marks the best reported shared-student score per column. Success rates are percentages, and $\uparrow$ indicates higher is better. CALVIN $P_{\geq k}$ measures completion of at least $k$ instructions, and average length equals $\sum_{k=1}^{5}P_{\geq k}/100$.}
\label{tab:main}
\footnotesize
\setlength{\tabcolsep}{2.5pt}
\begin{tabular*}{\linewidth}{@{\extracolsep{\fill}}lrrrrrrrrrrrr@{}}
\toprule
 & \multicolumn{5}{c}{LIBERO (\%) $\uparrow$} & MT50 $\uparrow$ & \multicolumn{6}{c}{CALVIN $\uparrow$} \\
\cmidrule(lr){2-6}\cmidrule(lr){8-13}
Method & All-40 & Spatial & Object & Goal & LIBERO-10 & SR & $P_{\geq1}$ & $P_{\geq2}$ & $P_{\geq3}$ & $P_{\geq4}$ & $P_{\geq5}$ & Length \\
\midrule
Domain specialists & 98.00 & 98.8 & 98.4 & 97.8 & 97.0 & 76.2 & 99.7 & 97.4 & 94.4 & 91.5 & 86.8 & 4.698 \\
Single-domain velocity OPD & 70.65 & 90.4 & 97.4 & 63.8 & 31.0 & 48.0 & 98.8 & 91.9 & 82.5 & 72.8 & 64.0 & 4.100 \\
\midrule
Shared SFT $\pi_{\theta_0}$ & 53.15 & 74.2 & 76.2 & 53.2 & 9.0 & 38.0 & 93.4 & 77.0 & 62.4 & 51.2 & 39.5 & 3.235 \\
Action loss (S/S) & 66.20 & \textbf{94.6} & 77.2 & 60.0 & 33.0 & 56.4 & 95.6 & 85.6 & 74.6 & 64.0 & 54.6 & 3.744 \\
Teacher-action CFM & 69.00 & 89.8 & 95.0 & 62.2 & 29.0 & 57.2 & \textbf{99.5} & 93.6 & 85.4 & 78.2 & 68.1 & 4.248 \\
Velocity loss (S/S, ours) & \textbf{75.70} & 94.2 & \textbf{98.4} & \textbf{63.8} & \textbf{46.4} & \textbf{60.6} & 99.0 & \textbf{94.7} & \textbf{88.4} & \textbf{81.2} & \textbf{72.9} & \textbf{4.362} \\
\bottomrule
\end{tabular*}
\end{table*}

\section{Problem Formulation}
\label{sec:prelim}

We consider consolidating $K$ frozen domain specialists $\{\pi_k^{\mathrm{T}}\}_{k=1}^{K}$---specifically $\pi_{\mathrm{LIBERO}}$, $\pi_{\mathrm{MetaWorld}}$, and $\pi_{\mathrm{CALVIN}}$---into one shared student policy $\pi_\theta$. At decision step $t$ in domain $k$, the student receives context $c_t=(o_t,g)$ containing visual and robot-state observations $o_t$ alongside language instruction $g$, generates a normalized action chunk $A$ of horizon $H$, executes the first $m$ actions ($m\leq H$), and replans from the next observation. While horizons and native action dimensions differ across domains, the student and specialist within each domain share aligned action representations and normalization statistics; all trainable parameters in $\pi_\theta$ are shared, leaving only input preprocessing and action decoding specific to each domain.

Starting from a joint demonstration-SFT checkpoint $\pi_{\theta_0}$, training queries the domain-matched teacher $\pi_k^{\mathrm{T}}$ on contexts drawn from the student's rollout distribution $d_k^\pi(c)$. During deployment, all three domains use the same policy weights and retain their own observation processing and action decoding.

Given context $c$, the student generates an action chunk by integrating a conditional velocity field over flow time $\tau\in[0,1]$ under the noise-to-action convention:
\begin{equation}
\begin{gathered}
 x_0=\epsilon\sim\Normal(0,I),\qquad A=x_1,\\
 \frac{dx_\tau}{d\tau}=v_\theta(x_\tau,\tau;c).
\end{gathered}
 \label{eq:ode}
\end{equation}
Flow time $\tau$ indexes internal synthesis at a fixed environment step $t$: $x_\tau$ is an intermediate action state distributed according to $q_{\theta,\tau}(x\mid c)$, so $d_k^\pi(c)$ and $q_{\theta,\tau}(x\mid c)$ together specify the environment and action-generation distributions from which supervision inputs are sampled. Integrating Equation~\eqref{eq:ode} with a deterministic Euler solver on a finite time grid yields the student and teacher action chunks $F_\theta(\epsilon,c)$ and $F_k^{\mathrm{T}}(\epsilon,c)$, where teacher $\pi_k^{\mathrm{T}}$ induces velocity field $v_k^{\mathrm{T}}$ and flow-state distribution $q_{k,\tau}^{\mathrm{T}}$.

\begin{figure*}[t]
\centering
\includegraphics[width=.94\linewidth]{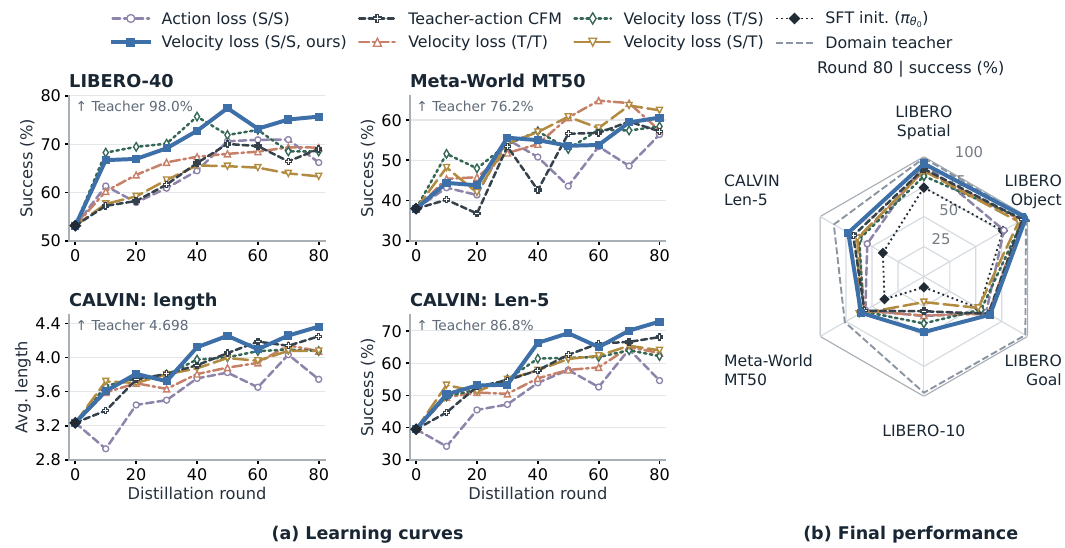}
\caption{\textbf{Multi-teacher consolidation across supervision designs.} (a) Unsmoothed learning curves for LIBERO-40, MT50, CALVIN average length, and CALVIN Len-5. (b) Round-80 success rates for the four LIBERO suites, MT50, and CALVIN Len-5. Black diamonds mark $\pi_{\theta_0}$ and gray dashed lines mark domain specialists.}
\label{fig:ab-evaluation}
\end{figure*}

\section{Multi-Teacher On-Policy Flow Distillation}
\label{sec:method}

As illustrated in Figure~\ref{fig:framework}, distillation alternates between closed-loop student collection, teacher-velocity labeling on student flow paths, and a weighted multi-domain update. Specifically, each round $r$ begins by freezing the rollout snapshot $\bar\theta_r=\sg[\theta_r]$ and running $\pi_{\bar\theta_r}$ in every domain, collecting complete successful and failed episodes so that on-policy contexts $c\sim d_k^{\pi_{\bar\theta_r}}$ reflect states encountered under student control.

At each collected context $c$, the student integrates its velocity field from noise $\epsilon$ through intermediate states $x\sim q_{\bar\theta_r,\tau}(\cdot\mid c)$ to generate the action chunk $A=F_{\bar\theta_r}(\epsilon,c)$. We query the specialist velocity field $v_k^{\mathrm{T}}(x,\tau;c)$ at every nonterminal node on this path; weighting domain $k$'s nonterminal solver nodes uniformly with $\rho(\tau\mid k,c)$ defines the multi-teacher \emph{Velocity Loss}:
\begin{equation}
\begin{aligned}
 \Loss_{\mathrm{vel},r}(\theta)&=
 \sum_{k=1}^{K}w_k\,
 \E_{\substack{c\sim d_k^{\pi_{\bar\theta_r}},\;\tau\sim\rho(\cdot\mid k,c)\\
               x\sim q_{\bar\theta_r,\tau}(\cdot\mid c)}}
 \\[2pt]
 &\quad\MSE\!\left(
 v_\theta(\sg[x],\tau;c),\sg[v_k^{\mathrm{T}}(x,\tau;c)]
 \right).
\end{aligned}
 \label{eq:opd}
\end{equation}
Here $w_k$ is the fixed loss weight for domain $k$, with $w_k\geq 0$ and $\sum_{k=1}^K w_k=1$. Within each domain, $\MSE$ averages squared errors across the action steps in a chunk and the active action coordinates before the domain losses are combined.

The student executes the first $m$ actions of $A$ before replanning. Equation~\eqref{eq:opd} matches velocities along the solver path that generated the full chunk. Whereas DiffusionOPD motivates velocity regression via the reverse KL between Gaussian transitions with shared covariance \citep{li2026diffusionopd}, holding both query states $\sg[x]$ and teacher targets $\sg[v_k^{\mathrm{T}}]$ fixed ensures that gradients pass only through the student's local velocity prediction at each nonterminal node. For a shared context, initial noise, and Euler grid, matching the teacher's velocities prior to the endpoint on either path yields identical integrated trajectories because every Euler update starts from the same state and takes the same step; once the velocity fields differ, however, the two paths can diverge, making the choice of supervised flow states an empirical question that we test in Section~\ref{sec:sampling-ablation}.

In each round, collecting $n$ episodes per domain, accumulating domain-weighted gradients of Equation~\eqref{eq:opd} across domains, and taking one clipped AdamW step completes an outer iteration (Algorithm~\ref{alg:training}; hyperparameters in Appendix~\ref{app:implementation}). As in VLA-OPD~\citep{zhong2026vlaopd}, the updated student collects fresh environment rollouts in the next round. The domain specialist supervises the new flow paths that the student uses to generate its actions. At deployment, only the student is needed.

\section{Experimental Evaluation}
\label{sec:experiments}

Table~\ref{tab:main} compares joint distillation with SFT, single-domain students, and two action-based baselines. Table~\ref{tab:sampling-sources} compares all four sampling configurations using the same velocity loss.


\begin{figure*}[t]
\centering
\includegraphics[width=.81\linewidth]{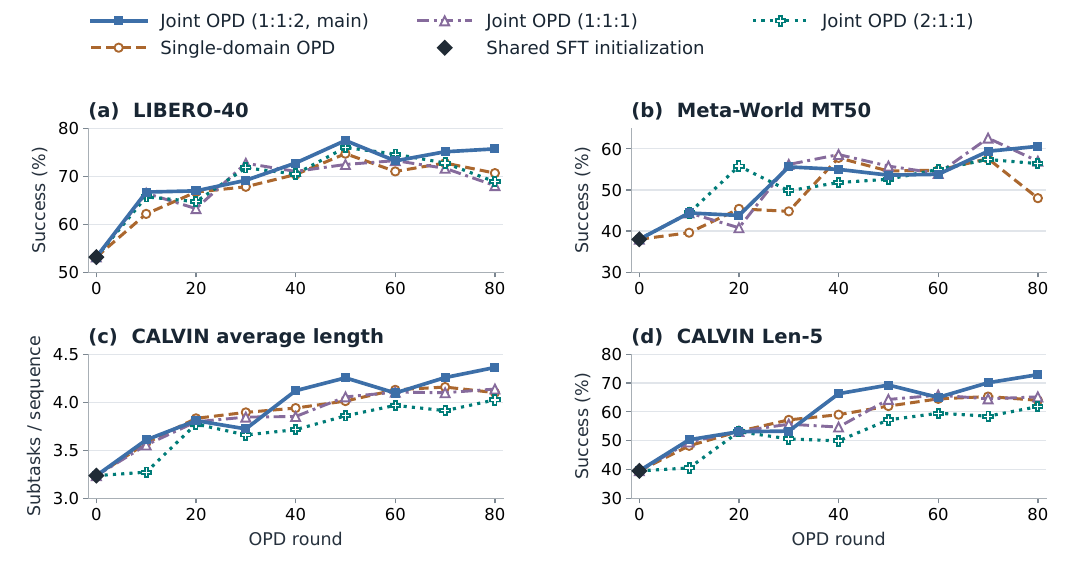}
\caption{\textbf{Joint and single-domain velocity distillation.} Joint curves show Velocity Loss MOPD with different weights (LIBERO, Meta-World, CALVIN). Orange circles mark single-domain students. All students start from $\pi_{\theta_0}$ (diamond), use S/S velocity supervision, and collect 64 episodes per active domain per round.}
\label{fig:single-domain-opd}
\end{figure*}

\subsection{Experimental Setup}

We evaluate across three manipulation benchmarks: LIBERO Spatial, Object, Goal, and LIBERO-10 (40 tasks total), Meta-World MT50, and CALVIN \citep{liu2023libero,yu2020metaworld,mees2022calvin}. Each evaluation checkpoint rolls out 2,000 LIBERO episodes, 500 Meta-World episodes, and 1,000 CALVIN sequences. LIBERO reports mean success across its four suites, Meta-World reports mean task success, and CALVIN reports both average completed length and five-instruction completion (Len-5). For both collection and evaluation, LIBERO and Meta-World use the same initial-state pools, whereas CALVIN collects rollouts in scenes A/B/C and evaluates on unseen scene D.

All distillation runs start from the same student, $\pi_{\theta_0}$, obtained after 14,000 joint SFT updates on demonstrations from LIBERO-40, MT50, and CALVIN scenes A/B/C. All methods share learning rates, trainable parameters (updating the action expert and language LoRA while freezing the vision backbone and teachers), within-domain normalization statistics, and evaluation-mode ODE solver settings (Appendix~\ref{app:implementation}).

Each optimizer update aggregates 64 episodes per active domain. The six joint configurations comprise two action-based baselines (Action Loss OPD and Teacher-action CFM) and four sampling combinations with the same velocity loss (S/S, S/T, T/S, and T/T). To emphasize CALVIN's five-instruction chains, all six use domain weights $(0.25,0.25,0.50)$ for LIBERO, MT50, and CALVIN, respectively ($1{:}1{:}2$). We evaluate every ten updates through round 80, matching the joint and single-domain students at 5,120 episodes per domain.

\subsection{Comparison Methods}
\label{sec:comparison-design}

Labels specify the loss and environment/flow-state sources (S: student; T: teacher), where S/S denotes student collection at both levels and Teacher-action CFM uses noise--action interpolants on student-visited states.

\label{sec:baselines}

\label{sec:endpoint}
The first baseline, \emph{Action Loss OPD}, matches terminal action chunks at student-visited contexts by integrating both student and teacher from shared initial noise $\epsilon$:
\begin{equation}
\begin{aligned}
 \Loss_{\mathrm{act},r}(\theta)&=
 \sum_{k=1}^{K}w_k\,
 \E_{\substack{c\sim d_k^{\pi_{\bar\theta_r}}\\
               \epsilon\sim\Normal(0,I)}}\\[2pt]
 &\quad\MSE\!\left(F_\theta(\epsilon,c),
 \sg[F_k^{\mathrm{T}}(\epsilon,c)]\right).
\end{aligned}
 \label{eq:endpoint}
\end{equation}
Here gradients backpropagate through the student's complete numerical ODE solve while holding the teacher endpoint fixed.

A second baseline, \emph{Teacher-action CFM}, avoids solver backpropagation by generating a frozen teacher chunk $A_k^{\mathrm{T}}=\sg[F_k^{\mathrm{T}}(\epsilon_T,c)]$ with \mbox{$\epsilon_T\sim\Normal(0,I)$} at each student-visited context and pairing $A_k^{\mathrm{T}}$ with independent noise $z\sim\Normal(0,I)$ at flow times $\tau\sim\rho_{\mathrm{CFM}}$ \citep{lipman2023flowmatching} (sampling details in Appendix~\ref{app:cfm-details}):
\begin{equation}
\begin{aligned}
 x_\tau^{\mathrm{CFM}}&=(1-\tau)z+\tau A_k^{\mathrm{T}},\\
 \Loss_{\mathrm{CFM},r}(\theta)&=\sum_{k=1}^{K}w_k\,\E_{c,A_k^{\mathrm{T}},z,\tau}\\[2pt]
 &\quad\MSE\!\left(v_\theta(x_\tau^{\mathrm{CFM}},\tau;c),
 A_k^{\mathrm{T}}-z\right).
\end{aligned}
\label{eq:teacher-action-cfm}
\end{equation}
In this baseline, gradients act locally on $v_\theta$, with the number of CFM samples per context matching each suite's solver-step count.

\label{sec:teacher-rollout}
Finally, to isolate sampling sources while keeping Equation~\eqref{eq:opd} fixed, we evaluate all four pairings of acting policy (first letter) and flow-state query path (second letter) in Table~\ref{tab:sampling-sources}. Whereas T/T and S/S reuse the acting policy's own flow trajectory, T/S and S/T integrate both policies from shared noise at the same context so one policy drives execution and the other supplies intermediate query states.

\begin{table*}[t]
\centering
\footnotesize
\caption{\textbf{Sources of environment contexts and internal flow states.} The first letter gives the acting policy and the second the query path (T for teacher, S for student). Schematic arrows at each open circle show the student velocity (filled blue arrowheads) and frozen teacher target (open red arrowheads). Results are at round 80, with success in percent and CALVIN length in instructions. Bold marks column maxima.}
\label{tab:sampling-sources}
\setlength{\tabcolsep}{2.2pt}
\begin{tabular*}{\linewidth}{@{\extracolsep{\fill}}lrrrrrrrrrrrr@{}}
\toprule
 & \multicolumn{5}{c}{LIBERO (\%) $\uparrow$} & MT50 $\uparrow$ & \multicolumn{6}{c}{CALVIN $\uparrow$} \\
\cmidrule(lr){2-6}\cmidrule(lr){8-13}
Method & All-40 & Spatial & Object & Goal & LIBERO-10 & SR & $P_{\geq1}$ & $P_{\geq2}$ & $P_{\geq3}$ & $P_{\geq4}$ & $P_{\geq5}$ & Length \\
\midrule
Velocity loss (T/T) & 69.30 & 88.4 & 94.0 & 62.2 & 32.6 & 57.4 & 98.1 & 92.2 & 81.3 & 72.6 & 63.2 & 4.074 \\
Velocity loss (T/S) & 68.50 & 83.8 & 95.8 & 55.2 & 39.2 & 58.4 & 97.9 & 91.9 & 82.1 & 73.5 & 62.2 & 4.076 \\
Velocity loss (S/T) & 63.35 & 88.2 & 91.4 & 52.4 & 21.4 & \textbf{62.4} & 97.2 & 91.0 & 81.7 & 73.7 & 64.0 & 4.076 \\
Velocity loss (S/S, ours) & \textbf{75.70} & \textbf{94.2} & \textbf{98.4} & \textbf{63.8} & \textbf{46.4} & 60.6 & \textbf{99.0} & \textbf{94.7} & \textbf{88.4} & \textbf{81.2} & \textbf{72.9} & \textbf{4.362} \\
\bottomrule
\end{tabular*}
\par\smallskip
\includegraphics[width=\linewidth]{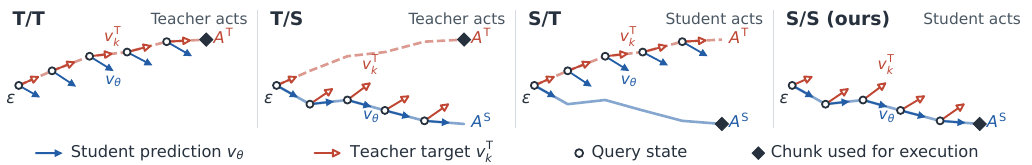}
\end{table*}

\subsection{Multi-Domain Capability Consolidation}
\label{sec:results}

Using a single set of shared weights, Velocity Loss MOPD raises student success to 75.70\% on LIBERO-40, 60.6\% on MT50, and 72.9\% on CALVIN Len-5 (Table~\ref{tab:main}), closing 50.3\%, 59.2\%, and 70.6\% of the respective gaps from the common SFT initialization to the domain specialists. Looking across individual LIBERO suites (learning curves in Appendix Figure~\ref{fig:ab-details}), the shared policy matches the specialist's 98.4\% success rate on Object (+21.2 points over Action Loss OPD) and reaches 46.4\% on LIBERO-10, outperforming both endpoint matching (33.0\%) and Teacher-action CFM (29.0\%).

Figure~\ref{fig:single-domain-opd} compares the shared student against three single-domain students trained independently under the same per-domain episode budget. With uniform weights at update 80, the shared student trails the single-domain student on LIBERO-40 by 2.60 points while leading on MT50 and CALVIN Len-5 by 9.0 and 1.2 points (Table~\ref{tab:main} and Appendix Table~\ref{tab:weight-ablation}). Under $1{:}1{:}2$ weights, it leads on all three benchmarks by 5.05, 12.6, and 8.9 points at update 80 (Table~\ref{tab:main}), and holds gains of 3.20 points on LIBERO-40, 4.47 on MT50, and 4.73 on CALVIN Len-5 when averaged over updates 60, 70, and 80 (Appendix Table~\ref{tab:joint-checkpoint-means}).

Among the three joint S/S runs in Figure~\ref{fig:single-domain-opd}, doubling CALVIN's loss weight ($1{:}1{:}2$) yields the highest LIBERO-40, MT50, and CALVIN Len-5 scores both at update 80 and on average over updates 60, 70, and 80. The largest difference across weights appears on LIBERO-10. At update 80, success reaches 46.4\% with double weight on CALVIN, 26.2\% with double weight on LIBERO ($2{:}1{:}1$), and 17.4\% with uniform weights ($1{:}1{:}1$) (Appendix Tables~\ref{tab:weight-ablation} and~\ref{tab:joint-checkpoint-means}).

\subsection{Comparing Supervision Targets}
\label{sec:supervision-targets}

Under identical student-driven collection (S/S), local velocity supervision exceeds terminal endpoint matching (Action Loss OPD) at every post-update evaluation (Figure~\ref{fig:ab-evaluation}(a)), finishing update 80 higher by 9.50 points on LIBERO-40, 4.20 on MT50, and 0.618 instructions on CALVIN. Action Loss OPD regresses between rounds 70 and 80 on LIBERO-40 (70.95\% to 66.20\%) and CALVIN Len-5 (64.3\% to 54.6\%; Appendix Table~\ref{tab:all-checkpoints}), and Velocity Loss MOPD still surpasses Action Loss OPD's round-70 peak by 4.75 and 8.60 points, respectively. On CALVIN, the margin widens steadily with chain length, from 3.4 points on $P_{\geq1}$ to 18.3 points on $P_{\geq5}$.

Velocity Loss MOPD also finishes ahead of Teacher-action CFM at update 80 by 6.70 points on LIBERO-40, 3.40 on MT50, and 4.80 on CALVIN Len-5 (Table~\ref{tab:main}). Across updates 60, 70, and 80, its mean lead persists on LIBERO-40 and CALVIN Len-5. MT50 means are nearly identical (57.93\% vs.\ 57.87\%, Appendix Tables~\ref{tab:all-checkpoints} and~\ref{tab:mixed-source-checkpoints}). Although both approaches roll out the student policy and supervise local velocities, Velocity Loss MOPD queries the teacher along the student's visited ODE path, while CFM trains on linear interpolants between independent noise and the teacher's terminal action (Equation~\eqref{eq:teacher-action-cfm}).

\subsection{Environment and Flow-State Sources}
\label{sec:sampling-ablation}

\label{sec:teacher-results}
Next, we hold the velocity objective fixed and test which policy should supply the environment contexts and internal flow states (Table~\ref{tab:sampling-sources}). At update 80, sampling from the student at both levels (S/S) outperforms teacher-rollout distillation (T/T) by 6.40 percentage points on LIBERO-40, 3.20 on MT50, and 9.70 on CALVIN Len-5 (+0.288 completed instructions per sequence on average), while remaining ahead of T/T on LIBERO-40 and both CALVIN metrics across all eight post-update evaluations (Figure~\ref{fig:ab-evaluation}(a)).

Varying the environment source reveals an interaction between the two sampling levels rather than an independent effect. On LIBERO-40 at update 80, switching from teacher to student rollouts raises success by 7.20 points with student flow states (S/S vs.\ T/S), yet lowers it by 5.95 points with teacher flow states (S/T vs.\ T/T), and the same reversal holds at updates 50, 60, and 70. On MT50, relative rankings shift over training, leaving T/T with higher mean success than S/S across updates 60, 70, and 80 (62.13\% vs.\ 57.93\%, Appendix Table~\ref{tab:all-checkpoints}).

A complementary asymmetry appears when changing the flow-state source. Averaged over updates 60, 70, and 80 on LIBERO-40 and CALVIN Len-5, replacing student flow states with teacher flow states on student rollouts (S/S to S/T) reduces success by 10.52 and 5.43 points, respectively, whereas making the same replacement on teacher rollouts (T/S to T/T) lowers success by only 0.95 and 0.43 points (Appendix Tables~\ref{tab:all-checkpoints} and~\ref{tab:mixed-source-checkpoints}). Over these checkpoints, MT50 instead favors teacher flow states on student rollouts (61.33\% vs.\ 57.93\%).

\section{Discussion and Limitations}
\label{sec:discussion}

Across our experiments, two-level alignment yields its largest gains on longer sequential tasks. On CALVIN, the success-rate advantage of S/S over endpoint action matching widens from 3.4 points on the first instruction to 18.3 points on five-instruction chains, and S/S also leads on LIBERO-10, where the policy executes ten actions open-loop before replanning---consistent with smaller chunk-level errors reducing error accumulation over longer horizons \citep{ross2011dagger}.

Although the consolidated policy shares all trainable transformer and flow-expert parameters, each domain retains its native action coordinates; gripper control, for instance, sits at coordinate 4 in Meta-World versus coordinate 7 in LIBERO and CALVIN. Even under equal rollout budgets, loss allocation shapes cross-domain trade-offs: doubling CALVIN's loss weight achieves the highest LIBERO-10 success, whereas uniform weights lead on LIBERO Spatial and Goal (Appendix Table~\ref{tab:weight-ablation}). Aligning action coordinates and velocity targets across domains and adapting domain loss weights or sampling proportions during training \citep{gao2026openmopd,sun2026d3mopd} are natural directions for future work.

Our empirical protocol also carries clear scope limits. Each configuration is evaluated from a single training run initialized at $\pi_{\theta_0}$ (64 episodes per domain per round), with late-stage means summarizing checkpoint performance within that run rather than seed-to-seed variance. In addition, LIBERO and Meta-World reuse the same initial-state pools for rollout collection and evaluation, whereas CALVIN holds out scene D from student training. Larger update budgets on LIBERO-10, multi-seed replications, and physical-robot deployment remain important next steps.

\section{Conclusion}
\label{sec:conclusion}

We introduced \emph{Velocity Loss MOPD}, a two-level multi-teacher framework that consolidates flow-based VLA specialists into one shared policy. By querying frozen specialists along student solver paths at student-visited observations, the method pairs local velocity regression with closed-loop control without backpropagating through the ODE solver. At update 80, it outperforms Action Loss OPD and Teacher-action CFM on all three benchmarks and leads all six joint runs in late-stage mean scores on LIBERO-40 and CALVIN. Overall, our method shows clear advantages on long-horizon tasks like LIBERO-10 and five-instruction CALVIN sequences, whereas its gains on shorter single-task MT50 rollouts are smaller and vary across checkpoints. On LIBERO-40 and CALVIN, our $2\times2$ ablation shows that querying student flow paths matters most when rollouts come from the student. Finally, under matched per-domain budgets, one shared policy remains competitive with single-domain distillation under uniform weights and surpasses all three single-domain students at update 80 and in late-stage means under $1{:}1{:}2$ weighting.

\section*{LLM Usage Statement}
We used OpenAI ChatGPT and Google Gemini to polish the manuscript writing. The authors take full responsibility for all analyses, citations, and conclusions.

\label{cvpr:references-start}
{\small
\bibliographystyle{ieeenat_fullname}
\bibliography{cvpr2027/references}

@inproceedings{kim2024openvla,
  title = {{OpenVLA}: An Open-Source Vision-Language-Action Model},
  author = {Kim, Moo Jin and Pertsch, Karl and Karamcheti, Siddharth and Xiao, Ted and Balakrishna, Ashwin and Nair, Suraj and Rafailov, Rafael and Foster, Ethan P. and Sanketi, Pannag R. and Vuong, Quan and Kollar, Thomas and Burchfiel, Benjamin and Tedrake, Russ and Sadigh, Dorsa and Levine, Sergey and Liang, Percy and Finn, Chelsea},
  year = {2025},
  url = {https://proceedings.mlr.press/v270/kim25c.html},
  booktitle = {Proceedings of the 8th Conference on Robot Learning},
  volume = {270},
  pages = {2679--2713},
  publisher = {PMLR}
}

@inproceedings{black2024pi0,
  title = {{$\pi_0$}: A Vision-Language-Action Flow Model for General Robot Control},
  author = {Black, Kevin and Brown, Noah and Driess, Danny and Esmail, Adnan and Equi, Michael Robert and Finn, Chelsea and Fusai, Niccolo and Groom, Lachy and Hausman, Karol and Ichter, Brian and Jakubczak, Szymon and Jones, Tim and Ke, Liyiming and Levine, Sergey and Li-Bell, Adrian and Mothukuri, Mohith and Nair, Suraj and Pertsch, Karl and Shi, Lucy Xiaoyang and Smith, Laura and Tanner, James and Vuong, Quan and Walling, Anna and Wang, Haohuan and Zhilinsky, Ury},
  year = {2025},
  doi = {10.15607/RSS.2025.XXI.010},
  url = {https://www.roboticsproceedings.org/rss21/p010.html},
  booktitle = {Proceedings of Robotics: Science and Systems}
}

@inproceedings{physicalintelligence2025pi05,
  title = {{$\pi_{0.5}$}: a Vision-Language-Action Model with Open-World Generalization},
  author = {Black, Kevin and Brown, Noah and Darpinian, James and Dhabalia, Karan and Driess, Danny and Esmail, Adnan and Equi, Michael Robert and Finn, Chelsea and Fusai, Niccolo and Galliker, Manuel Y. and Ghosh, Dibya and Groom, Lachy and Hausman, Karol and Ichter, Brian and Jakubczak, Szymon and Jones, Tim and Ke, Liyiming and LeBlanc, Devin and Levine, Sergey and Li-Bell, Adrian and Mothukuri, Mohith and Nair, Suraj and Pertsch, Karl and Ren, Allen Z. and Shi, Lucy Xiaoyang and Smith, Laura and Springenberg, Jost Tobias and Stachowicz, Kyle and Tanner, James and Vuong, Quan and Walke, Homer and Walling, Anna and Wang, Haohuan and Yu, Lili and Zhilinsky, Ury},
  year = {2025},
  url = {https://proceedings.mlr.press/v305/black25a.html},
  booktitle = {Proceedings of the 9th Conference on Robot Learning},
  volume = {305},
  pages = {17--40},
  publisher = {PMLR}
}

@inproceedings{chi2023diffusionpolicy,
  title = {Diffusion Policy: Visuomotor Policy Learning via Action Diffusion},
  author = {Chi, Cheng and Feng, Siyuan and Du, Yilun and Xu, Zhenjia and Cousineau, Eric and Burchfiel, Benjamin and Song, Shuran},
  booktitle = {Proceedings of Robotics: Science and Systems},
  year = {2023},
  doi = {10.15607/RSS.2023.XIX.026},
  url = {https://www.roboticsproceedings.org/rss19/p026.html}
}

@inproceedings{lipman2023flowmatching,
  title = {Flow Matching for Generative Modeling},
  author = {Lipman, Yaron and Chen, Ricky T. Q. and Ben-Hamu, Heli and Nickel, Maximilian and Le, Matthew},
  booktitle = {The Eleventh International Conference on Learning Representations},
  year = {2023},
  url = {https://openreview.net/forum?id=PqvMRDCJT9t}
}

@inproceedings{fu2025mergevla,
  title = {{MergeVLA}: Cross-Skill Model Merging Toward a Generalist Vision-Language-Action Agent},
  author = {Fu, Yuxia and Zhang, Zhizhen and Zhang, Yuqi and Wang, Zijian and Huang, Zi and Luo, Yadan},
  year = {2026},
  url = {https://openaccess.thecvf.com/content/CVPR2026/html/Fu_MergeVLA_Cross-Skill_Model_Merging_Toward_a_Generalist_Vision-Language-Action_Agent_CVPR_2026_paper.html},
  booktitle = {Proceedings of the IEEE/CVF Conference on Computer Vision and Pattern Recognition},
  pages = {22335--22347}
}

@article{fang2025dualvla,
  title = {{DualVLA}: Building a Generalizable Embodied Agent via Partial Decoupling of Reasoning and Action},
  author = {Fang, Zhen and Liu, Zhuoyang and Liu, Jiaming and Chen, Hao and Zeng, Yu and Huang, Shiting and Chen, Zehui and Chen, Lin and Zhang, Shanghang and Zhao, Feng},
  journal = {arXiv preprint arXiv:2511.22134},
  year = {2025},
  doi = {10.48550/arXiv.2511.22134},
  url = {https://arxiv.org/abs/2511.22134}
}

@inproceedings{xiao2026pld,
  title = {Self-Improving Vision-Language-Action Models with Data Generation via Residual {RL}},
  author = {Xiao, Wenli and Lin, Haotian and Peng, Andy and Xue, Haoru and He, Tairan and Xie, Yuqi and Hu, Fengyuan and Wu, Jimmy and Luo, Zhengyi and Fan, Linxi {``Jim''} and Shi, Guanya and Zhu, Yuke},
  booktitle = {The Fourteenth International Conference on Learning Representations},
  year = {2026},
  url = {https://proceedings.iclr.cc/paper_files/paper/2026/hash/163a875ab0c2631ed30b7435c4c9b8f1-Abstract-Conference.html}
}

@inproceedings{xu2025rldg,
  title = {{RLDG}: Robotic Generalist Policy Distillation via Reinforcement Learning},
  author = {Xu, Charles and Li, Qiyang and Luo, Jianlan and Levine, Sergey},
  booktitle = {Proceedings of Robotics: Science and Systems},
  year = {2025},
  doi = {10.15607/RSS.2025.XXI.028},
  url = {https://www.roboticsproceedings.org/rss21/p028.html}
}

@inproceedings{xu2025mmnav,
  title = {{MM-Nav}: Multi-View {VLA} Model for Robust Visual Navigation via Multi-Expert Learning},
  author = {Xu, Tianyu and Chen, Jiawei and Zhang, Jiazhao and Zhang, Wenyao and Qi, Zekun and Li, Minghan and Liu, Jiahang and Yue, Lu and Zhang, Zhizheng and Wang, He},
  year = {2026},
  url = {https://pku-epic.github.io/MM-Nav-Web/},
  booktitle = {European Conference on Computer Vision}
}

@inproceedings{ross2011dagger,
  title     = {A Reduction of Imitation Learning and Structured Prediction to No-Regret Online Learning},
  author    = {Ross, Stephane and Gordon, Geoffrey and Bagnell, Drew},
  booktitle = {Proceedings of the Fourteenth International Conference on Artificial Intelligence and Statistics},
  volume    = {15},
  series    = {Proceedings of Machine Learning Research},
  pages     = {627--635},
  publisher = {PMLR},
  year      = {2011},
  url       = {https://proceedings.mlr.press/v15/ross11a.html}
}

@inproceedings{rusu2016policy,
  title     = {Policy Distillation},
  author    = {Rusu, Andrei A. and Gomez Colmenarejo, Sergio and Gulcehre, Caglar and Desjardins, Guillaume and Kirkpatrick, James and Pascanu, Razvan and Mnih, Volodymyr and Kavukcuoglu, Koray and Hadsell, Raia},
  booktitle = {International Conference on Learning Representations},
  year      = {2016},
  url       = {https://arxiv.org/abs/1511.06295}
}

@inproceedings{czarnecki2019distilling,
  title     = {Distilling Policy Distillation},
  author    = {Czarnecki, Wojciech M. and Pascanu, Razvan and Osindero, Simon and Jayakumar, Siddhant and Swirszcz, Grzegorz and Jaderberg, Max},
  booktitle = {Proceedings of the Twenty-Second International Conference on Artificial Intelligence and Statistics},
  volume    = {89},
  series    = {Proceedings of Machine Learning Research},
  pages     = {1331--1340},
  publisher = {PMLR},
  year      = {2019},
  url       = {https://proceedings.mlr.press/v89/czarnecki19a.html}
}

@inproceedings{agarwal2024gkd,
  title     = {On-Policy Distillation of Language Models: Learning from Self-Generated Mistakes},
  author    = {Agarwal, Rishabh and Vieillard, Nino and Zhou, Yongchao and Stanczyk, Piotr and Ramos Garea, Sabela and Geist, Matthieu and Bachem, Olivier},
  booktitle = {International Conference on Learning Representations},
  year      = {2024},
  url       = {https://proceedings.iclr.cc/paper_files/paper/2024/hash/5be69a584901a26c521c2b51e40a4c20-Abstract-Conference.html}
}

@article{zhong2026vlaopd,
  title   = {{VLA-OPD}: Bridging Offline {SFT} and Online {RL} for Vision-Language-Action Models via On-Policy Distillation},
  author  = {Zhong, Zhide and Yan, Haodong and Li, Junfeng and He, Junjie and Zhang, Tianran and Li, Haoang},
  journal = {arXiv preprint arXiv:2603.26666},
  year    = {2026},
  doi     = {10.48550/arXiv.2603.26666},
  url     = {https://arxiv.org/abs/2603.26666}
}

@article{li2026diffusionopd,
  title   = {{DiffusionOPD}: A Unified Perspective of On-Policy Distillation in Diffusion Models},
  author  = {Li, Quanhao and Yu, Junqiu and Jiang, Kaixun and Wei, Yujie and Xing, Zhen and Li, Pandeng and Chu, Ruihang and Zhang, Shiwei and Liu, Yu and Wu, Zuxuan},
  journal = {arXiv preprint arXiv:2605.15055},
  year    = {2026},
  doi     = {10.48550/arXiv.2605.15055},
  url     = {https://arxiv.org/abs/2605.15055v2}
}

@article{ma2026mopd,
  title   = {{MOPD}: Multi-Teacher On-Policy Distillation for Capability Integration in {LLM} Post-Training},
  author  = {Ma, Wenhan and Wei, Jianyu and Zhao, Liang and Zhang, Hailin and Xiao, Bangjun and Li, Lei and Yang, Qibin and Gao, Bofei and Wang, Yudong and Li, Rang and Dong, Jinhao and Sui, Zhifang and Luo, Fuli},
  journal = {arXiv preprint arXiv:2606.30406},
  year    = {2026},
  doi     = {10.48550/arXiv.2606.30406},
  url     = {https://arxiv.org/abs/2606.30406}
}

@article{gao2026openmopd,
  title   = {{Open-MOPD}: Diagnosing and Fixing Capability Imbalance in Multi-Teacher On-Policy Distillation},
  author  = {Gao, Huan-ang and Chi, Haohan and Yan, Yong and Feng, Shiyuan and Wu, Hanlin and Jiang, Zheng and He, Bingxiang and Ma, Wei-Ying and Zhang, Ya-Qin and Zhou, Hao},
  journal = {arXiv preprint arXiv:2608.19098},
  year    = {2026},
  doi     = {10.48550/arXiv.2608.19098},
  url     = {https://arxiv.org/abs/2608.19098}
}

@article{sun2026d3mopd,
  title   = {{D}$^{3}$-{MOPD}: Dynamic Domain {ScheDuling} for Efficient Multi-Teacher Distillation},
  author  = {Sun, Zechen and Zhang, Zhiwei and Zhao, Fei and Li, Juntao and Chuan, Mu and Deng, Huayu and Zhan, Guojian and Chen, Wenliang and Hu, Yao and Zhang, Min},
  journal = {arXiv preprint arXiv:2608.24987},
  year    = {2026},
  doi     = {10.48550/arXiv.2608.24987},
  url     = {https://arxiv.org/abs/2608.24987}
}

@inproceedings{liu2023libero,
  title = {{LIBERO}: Benchmarking Knowledge Transfer for Lifelong Robot Learning},
  author = {Liu, Bo and Zhu, Yifeng and Gao, Chongkai and Feng, Yihao and Liu, Qiang and Zhu, Yuke and Stone, Peter},
  booktitle = {Advances in Neural Information Processing Systems},
  year = {2023},
  url = {https://proceedings.neurips.cc/paper_files/paper/2023/hash/8c3c666820ea055a77726d66fc7d447f-Abstract-Datasets_and_Benchmarks.html},
  volume = {36},
  pages = {44776--44791},
  publisher = {Curran Associates, Inc.},
  doi = {10.52202/075280-1939}
}

@inproceedings{yu2020metaworld,
  title = {{Meta-World}: A Benchmark and Evaluation for Multi-Task and Meta Reinforcement Learning},
  author = {Yu, Tianhe and Quillen, Deirdre and He, Zhanpeng and Julian, Ryan and Hausman, Karol and Finn, Chelsea and Levine, Sergey},
  booktitle = {Proceedings of the Conference on Robot Learning},
  volume = {100},
  series = {Proceedings of Machine Learning Research},
  pages = {1094--1100},
  year = {2020},
  publisher = {PMLR},
  url = {https://proceedings.mlr.press/v100/yu20a.html}
}

@article{mees2022calvin,
  title = {{CALVIN}: A Benchmark for Language-Conditioned Policy Learning for Long-Horizon Robot Manipulation Tasks},
  author = {Mees, Oier and Hermann, Lukas and Rosete-Beas, Erick and Burgard, Wolfram},
  journal = {IEEE Robotics and Automation Letters},
  volume = {7},
  number = {3},
  pages = {7327--7334},
  year = {2022},
  url = {https://arxiv.org/abs/2112.03227},
  doi = {10.1109/LRA.2022.3180108}
}

@article{chen2025pirl,
  title = {{$\pi_{\mathrm{RL}}$}: Online {RL} Fine-tuning for Flow-based Vision-Language-Action Models},
  author = {Chen, Kang and Liu, Zhihao and Zhang, Tonghe and Guo, Zhen and Xu, Si and Lin, Hao and Zang, Hongzhi and Li, Xiang and Wei, Bingwen and Zhou, Jiakai and Zhang, Quanlu and Yu, Zhaofei and Fan, Guoliang and Huang, Tiejun and Wang, Yu and Yu, Chao},
  journal = {arXiv preprint arXiv:2510.25889},
  year = {2025},
  doi = {10.48550/arXiv.2510.25889},
  url = {https://arxiv.org/abs/2510.25889v3}
}

@article{oh2026fastopd,
  title = {{FastOPD}: On-Policy Distillation for Lightweight {VLA} Deployment},
  author = {Oh, Yoojin and Kim, Jeongsol and Seo, Yeonwoo and Park, Jangho and Jin, Seonghyun and Park, Sunwoo and Kim, Youngmin and Jun, Youngjun and Choi, Kyumin and Ye, Jong Chul},
  journal = {arXiv preprint arXiv:2610.02832},
  year = {2026},
  url = {https://arxiv.org/abs/2610.02832v1}
}

@article{deshmukh2026multitaskcfm,
  title = {Distillation for Efficient Multitask Manipulation Policies via Conditional Flow Matching},
  author = {Deshmukh, Shreya and Mahdi, Imen and Heppert, Nick and Valada, Abhinav},
  journal = {arXiv preprint arXiv:2609.28107},
  year = {2026},
  url = {https://arxiv.org/abs/2609.28107v2}
}

@article{jeon2026shallowpi,
  title = {Shallow-{$\pi$}: Knowledge Distillation for Flow-based {VLAs}},
  author = {Jeon, Boseong and Choi, Yunho and Kim, Taehan},
  journal = {arXiv preprint arXiv:2601.20262},
  year = {2026},
  url = {https://arxiv.org/abs/2601.20262v1}
}

@article{yang2026wamopd,
  title = {{WAM-OPD}: Joint Video-Action Supervision for World Action Model Post-Training with On-Policy Distillation},
  author = {Yang, Liuhaichen and Zhong, Zhengyang and Zhu, Hanshang and Bai, Ningwei and Yin, Qichen and Han, Zhi and Qin, Jiarui and Jiang, Zhuang and Sheng, Chenchao and Ma, Hanbo and Liu, Junkai and Sun, Junkai and Lyu, Dongcheng and Liu, Bo and Dong, Yi and Tang, Zezhi},
  journal = {arXiv preprint arXiv:2608.22364},
  year = {2026},
  url = {https://arxiv.org/abs/2608.22364v2}
}

@article{luan2026snapflow,
  title = {{SnapFlow}: One-Step Action Generation for Flow-Matching {VLAs} via Progressive Self-Distillation},
  author = {Luan, Wuyang and Li, Junhui and Zhao, Weiguang and Zhang, Wenjian and Wu, Tieru and Ma, Rui},
  journal = {arXiv preprint arXiv:2604.05656},
  year = {2026},
  url = {https://arxiv.org/abs/2604.05656v1}
}

@article{jun2026driftopd,
  title = {{DriftOPD}: Sequence-Level Reverse-{KL} Distillation for One-Step {VLA} Policies},
  author = {Jun, Youngjun and Choi, Kyumin and Kim, Youngmin and Jin, Seonghyun and Park, Sunwoo and Park, Jangho and Ye, Jong Chul},
  journal = {arXiv preprint arXiv:2610.00317},
  year = {2026},
  url = {https://arxiv.org/abs/2610.00317v1}
}

@misc{ren2026uaopd,
  title = {When to Trust Your Teacher: Uncertainty-Aware On-Policy Distillation for Embodied Policies},
  author = {Ren, Haoxiang and Sheng, Yifei and Zhang, Zhilong and Xu, Runjie and Lu, Bowen and Lin, Haoxin and Yu, Yang},
  year = {2026},
  howpublished = {RSS Workshop on Diffusion for Robot Learning},
  url = {https://openreview.net/pdf?id=hHUZi8oHfo}
}

@article{liu2026wamopd,
  title = {{WAM-OPD}: Sharpening World Action Models via On-Policy Distillation},
  author = {Liu, Panjun and Lei, Xiaohan and Zhang, Shiqi and Wang, Yikun and Zhang, Yongxin and Hu, Mingyi and Sun, Shida and Shou, Jiateng and Zhou, Wengang and Deng, Jiajun and Xiong, Zhiwei},
  journal = {arXiv preprint arXiv:2609.34250},
  year = {2026},
  doi = {10.48550/arXiv.2609.34250},
  url = {https://arxiv.org/abs/2609.34250v1}
}

@article{xie2026rethinking,
  title = {Rethinking Self-Distillation for Multi-Teacher Capability Merging},
  author = {Xie, Roy and Friedman, Dan and Nan, Feng and Huang, Yukun and Xu, Zhichao and Zhang, Chengjiu and Xu, Jun and Faruqui, Manaal and Rathod, Vivek and Dhingra, Bhuwan},
  journal = {arXiv preprint arXiv:2610.04272},
  year = {2026},
  doi = {10.48550/arXiv.2610.04272},
  url = {https://arxiv.org/abs/2610.04272v1}
}

@article{li2026dnmopd,
  title = {Beyond Teacher Assignment: Domain-Normalized Multi-Teacher On-Policy Distillation},
  author = {Li, Xin and Jiang, Hao and Gao, Xin and Wang, Annan and Xie, Yuchen and Guo, Jinghao and Qu, Xingwei and Zhang, Yichi and Yuen, Chau},
  journal = {arXiv preprint arXiv:2609.35347},
  year = {2026},
  doi = {10.48550/arXiv.2609.35347},
  url = {https://arxiv.org/abs/2609.35347v1}
}

@article{zhu2026upmopd,
  title = {{UP-MOPD}: Update Projection in Multi-Teacher On-Policy Distillation},
  author = {Zhu, Taojie and Jin, Jing and Xia, Yuan and Ding, Chenyang and He, Qunshan and Xia, Wanke and Sun, Tao and Chen, Yan and Wang, Jian and Gu, Jinjie and Feng, Tao},
  journal = {arXiv preprint arXiv:2610.08398},
  year = {2026},
  doi = {10.48550/arXiv.2610.08398},
  url = {https://arxiv.org/abs/2610.08398v1}
}

@article{fang2026flowopd,
  title = {{Flow-OPD}: On-Policy Distillation for Flow Matching Models},
  author = {Fang, Zhen and Huang, Wenxuan and Zeng, Yu and Zhao, Yiming and Chen, Shuang and Feng, Kaituo and Lin, Yunlong and Chen, Lin and Chen, Zehui and Cao, Shaosheng and Zhao, Feng},
  journal = {arXiv preprint arXiv:2605.08063},
  year = {2026},
  url = {https://arxiv.org/abs/2605.08063v5}
}

@article{ruan2026pivotq,
  title = {Beyond In-Distribution Preservation: Recovering Generalization in Quantized {VLAs} via Vulnerability-Oriented Tuning},
  author = {Ruan, Shen and Gao, Wenchang and Wang, Jin and Liu, Siao and {Zhaxizhuoma} and Ren, Dongchun and Zheng, Xin},
  journal = {arXiv preprint arXiv:2610.05745},
  year = {2026},
  url = {https://arxiv.org/abs/2610.05745v2}
}

@article{li2026flowmapopd,
  title = {{FlowMap-OPD}: Rollout--Kernel Separation for On-Policy Distillation of Few-Step Flow-Map Generators},
  author = {Li, Zhiqi and Zhu, Bo},
  journal = {arXiv preprint arXiv:2609.37851},
  year = {2026},
  url = {https://arxiv.org/abs/2609.37851v1}
}

@inproceedings{gu2024minillm,
  title = {{MiniLLM}: Knowledge Distillation of Large Language Models},
  author = {Gu, Yuxian and Dong, Li and Wei, Furu and Huang, Minlie},
  booktitle = {International Conference on Learning Representations},
  year = {2024},
  url = {https://proceedings.iclr.cc/paper_files/paper/2024/hash/8ac015d409635f196f9e3e9dcfb9a94e-Abstract-Conference.html}
}

@misc{lu2025onpolicydistillation,
  title = {On-Policy Distillation},
  author = {Lu, Kevin and {Thinking Machines Lab}},
  year = {2025},
  month = oct,
  howpublished = {Thinking Machines Lab: Connectionism},
  note = {Research blog, October 27},
  doi = {10.64434/tml.20251026},
  url = {https://thinkingmachines.ai/blog/on-policy-distillation/}
}

@article{lu2026tilde,
  title = {{TILDE}: Teacher-in-the-Loop Distillation from Embodied Foundation Models},
  author = {Lu, Yuxiang and Liu, Yantong and Hou, Jinghua and Sun, Xueli and Han, Ruihua and Liu, Zhe and Zhao, Hengshuang},
  journal = {arXiv preprint arXiv:2610.25376},
  year = {2026},
  url = {https://arxiv.org/abs/2610.25376v1}
}

@article{jiang2026repair,
  title = {Beyond Policy Patches: Distilling {RL} Specialists into Vision-Language-Action Policies},
  author = {Jiang, Zhennan and Xie, Tingying and Ji, Shilong and Yu, Shu'ang and Zhou, Tianxing and Chen, Yuhui and Gao, Jiahuan and Xie, Kaizhi and Xu, Yuanfan and Gao, Feng and Wang, Yu and Li, Haoran and Zhao, Dongbin and Yu, Chao},
  journal = {arXiv preprint arXiv:2611.02777},
  year = {2026},
  url = {https://arxiv.org/abs/2611.02777v1}
}
}

\clearpage
\makeatletter\let\Hy@backout\@gobble\makeatother
\maketitlesupplementary
\appendix

\section{Implementation Details}
\label{app:implementation}

During distillation, the student and each domain specialist share that specialist's action-normalization statistics. The frozen pretrained \pihalf\ specialists are:
\begin{itemize}
\setlength\itemsep{0pt}
\item LIBERO: \texttt{pi05\_libero};
\item Meta-World: \texttt{RLinf-Pi05-MetaWorld-RL-FlowSDE};
\item CALVIN: \texttt{RLinf-Pi05-CALVIN-ABC-D-RL-FlowSDE}.
\end{itemize}
For CALVIN, we use the \texttt{InternData-Calvin\_ABC} normalization statistics supplied with the specialist. Student SFT and distillation use scenes A/B/C, and we evaluate the student in scene D.

\paragraph{Shared SFT initialization.}
The shared initialization $\pi_{\theta_0}$ is obtained after 14,000 joint SFT updates of \pihalf\ base (seed 0) on demonstrations from LIBERO-40 (1,880 trajectories), MT50 (2,350), and CALVIN scenes A/B/C (15,178). Each update draws 64 action windows per domain, sampled uniformly across tasks for LIBERO and MT50 and balanced by skill for CALVIN. Training optimizes the action expert (peak learning rate $2.5\times10^{-5}$) and rank-64 language LoRA ($5\times10^{-5}$) with CFM over valid time steps and native action coordinates while keeping the vision backbone frozen; both parameter groups use 1,500 warmup steps followed by cosine decay (Figure~\ref{fig:sft-training}).

\begin{strip}
\begin{minipage}{\textwidth}
\captionsetup{hypcap=false}
\centering
\captionof{table}{Settings for the six joint distillation configurations in the main comparison.}
\footnotesize
\begin{tabularx}{\linewidth}{@{}lX@{}}
\toprule
Parameter & Setting \\
\midrule
Shared student & Joint SFT step 14,000; shared parameters across three domains \\
Trainable parameters & 693,422,112 expert parameters and 78,446,592 LoRA parameters \\
LoRA & Rank 64, alpha 128, dropout 0; vision backbone frozen \\
Optimizer & AdamW, betas $(0.9,0.95)$, epsilon $10^{-8}$, weight decay $10^{-10}$ \\
Learning rates & Expert $2.5\times10^{-5}$; LoRA $5\times10^{-5}$; constant, no warmup \\
Update settings & Per-step microbatch 8 (80\,GB) or 4 (40\,GB); full-round gradient accumulation; clipping 1.0 \\
Domain weights & LIBERO / Meta-World / CALVIN: $0.25/0.25/0.50$ \\
Collection & $n=64$ episodes per domain; 192 total per round; one optimizer update \\
Collection order & LIBERO $\rightarrow$ Meta-World $\rightarrow$ CALVIN \\
Normalization & Shared teacher statistics within each domain; native action mask \\
Generation & Configured \texttt{eval\_flow\_ode}; shared batched noise-seed convention \\
Training seed & 42 across all distillation configurations \\
\bottomrule
\end{tabularx}
\end{minipage}
\end{strip}

\paragraph{Action interface.}
All comparisons use the same action interface within each suite. The chunk length $H$, number of actions executed before replanning $m$, and number of solver steps $N$ are $(H,m,N)=(5,5,5)$ for Meta-World and CALVIN, $(10,5,3)$ for LIBERO-Spatial, $(10,5,5)$ for Object and Goal, and $(10,10,5)$ for LIBERO-10. Native action dimensions are 4 for Meta-World and 7 for LIBERO and CALVIN. For teacher-velocity matching, we average losses over the nonterminal nodes of each flow path.

\paragraph{Hardware and collection budgets.}
Runs use four NVIDIA H800 80\,GB GPUs (microbatch 8) or eight A100 40\,GB GPUs (microbatch 4), accumulating gradients across each round's collected contexts ($n=64$ episodes per active domain) before one clipped AdamW step. Over 80 updates, each joint run collects 15,360 episodes (5,120 per domain), matching each single-domain student's 5,120 episodes. All distillation runs use training seed 42. Evaluation rollouts are collected independently of training rollouts.

\paragraph{Sampling variants.}
T/T, T/S, and S/T retain the same velocity MSE objective, domain weights, and optimizer settings. In T/S and S/T, both policies integrate from the same initial noise; the selected environment actor executes actions, and teacher velocities supervise the selected flow path.

\paragraph{Teacher-action CFM settings.}
\label{app:cfm-details}
Teacher and student ODEs start from the same noise at each context. Training pairs independent Gaussian noise with OpenPI's Beta$(1.5,1)$ time samples (three per context on LIBERO-Spatial, five elsewhere), averaging the losses. In OpenPI, $\widetilde\tau=0$ denotes action and $\widetilde\tau=1$ noise. Setting $\tau=1-\widetilde\tau$ reverses the velocity sign, giving Equation~\eqref{eq:teacher-action-cfm}.

\clearpage
\section{Additional Experimental Results}
\label{app:reporting}

\subsection{Checkpoint Evaluations}

Tables~\ref{tab:all-checkpoints} and~\ref{tab:mixed-source-checkpoints} report checkpoint evaluations for the six joint distillation configurations (learning curves in Figure~\ref{fig:ab-evaluation}).

\begin{strip}
\begin{minipage}{\textwidth}
\captionsetup{hypcap=false}
\centering
\captionof{table}{Checkpoint evaluations through round 80. Labels specify loss and environment/flow-state sources (S: student; T: teacher). Velocity loss (S/S) is our full method. Round 0 repeats the common $\pi_{\theta_0}$ evaluation. LIBERO and MT50 entries are success percentages. CALVIN entries report average completed length / Len-5 success rate (\%), where Len-5 means completing all five instructions. The mean row averages updates 60, 70, and 80.}
\label{tab:all-checkpoints}
\footnotesize
\setlength{\tabcolsep}{2.5pt}
\begin{tabular}{@{}rrrrrrrrrr@{}}
\toprule
Round & \multicolumn{3}{c}{LIBERO-40 (\%)} & \multicolumn{3}{c}{MT50 (\%)} & \multicolumn{3}{c}{CALVIN (length / Len-5 \%)} \\
\cmidrule(lr){2-4}\cmidrule(lr){5-7}\cmidrule(lr){8-10}
 & \shortstack{Action loss\\(S/S)} & \shortstack{Velocity loss\\(S/S)} & \shortstack{Velocity loss\\(T/T)} & \shortstack{Action loss\\(S/S)} & \shortstack{Velocity loss\\(S/S)} & \shortstack{Velocity loss\\(T/T)} & \shortstack{Action loss\\(S/S)} & \shortstack{Velocity loss\\(S/S)} & \shortstack{Velocity loss\\(T/T)} \\
\midrule
0 & 53.15 & 53.15 & 53.15 & 38.0 & 38.0 & 38.0 & 3.235/39.5 & 3.235/39.5 & 3.235/39.5 \\
10 & 61.35 & 66.70 & 60.25 & 43.2 & 44.4 & 45.4 & 2.930/34.2 & 3.607/50.3 & 3.590/49.4 \\
20 & 57.95 & 66.95 & 63.65 & 41.4 & 43.8 & 45.8 & 3.444/45.5 & 3.808/53.1 & 3.698/50.9 \\
30 & 60.90 & 69.15 & 66.25 & 55.2 & 55.6 & 51.8 & 3.500/47.2 & 3.722/53.3 & 3.632/50.5 \\
40 & 64.50 & 72.75 & 67.40 & 50.8 & 55.0 & 54.0 & 3.755/53.9 & 4.120/66.3 & 3.806/55.3 \\
50 & 70.60 & 77.45 & 68.00 & 43.6 & 53.6 & 60.6 & 3.823/57.9 & 4.254/69.3 & 3.881/57.9 \\
60 & 70.90 & 73.20 & 68.45 & 53.4 & 53.8 & 64.8 & 3.651/52.6 & 4.096/65.0 & 3.937/58.7 \\
70 & 70.95 & 75.10 & 69.35 & 48.6 & 59.4 & 64.2 & 4.033/64.3 & 4.257/70.1 & 4.143/64.9 \\
80 & 66.20 & 75.70 & 69.30 & 56.4 & 60.6 & 57.4 & 3.744/54.6 & 4.362/72.9 & 4.074/63.2 \\
\midrule
Mean & 69.35 & 74.67 & 69.03 & 52.80 & 57.93 & 62.13 & 3.809 / 57.17 & 4.238 / 69.33 & 4.051 / 62.27 \\
\bottomrule
\end{tabular}
\end{minipage}
\end{strip}

\begin{strip}
\begin{minipage}{\textwidth}
\captionsetup{hypcap=false}
\centering
\footnotesize
\captionof{table}{\textbf{Additional checkpoint evaluations.} T/S and S/T denote velocity-loss environment/flow-state sources (S: student; T: teacher). CFM denotes Teacher-action CFM. LIBERO and MT50 entries are success rates (\%); CALVIN reports average length / Len-5 success rate (\%). The mean row averages updates 60, 70, and 80.}
\label{tab:mixed-source-checkpoints}
\setlength{\tabcolsep}{3pt}
\begin{tabular*}{\linewidth}{@{\extracolsep{\fill}}rrrrrrrrrr@{}}
\toprule
 & \multicolumn{3}{c}{LIBERO-40 (\%)} & \multicolumn{3}{c}{MT50 (\%)} & \multicolumn{3}{c}{CALVIN (length / Len-5 \%)} \\
\cmidrule(lr){2-4}\cmidrule(lr){5-7}\cmidrule(lr){8-10}
Round & T/S & S/T & CFM & T/S & S/T & CFM & T/S & S/T & CFM \\
\midrule
10 & 68.25 & 57.65 & 57.25 & 51.6 & 48.2 & 40.2 & 3.666 / 49.7 & 3.720 / 53.2 & 3.377 / 44.7 \\
20 & 69.45 & 59.20 & 58.25 & 48.0 & 42.0 & 36.8 & 3.754 / 51.8 & 3.702 / 51.2 & 3.740 / 52.7 \\
30 & 70.10 & 62.60 & 61.65 & 54.0 & 54.2 & 53.4 & 3.800 / 55.0 & 3.801 / 55.0 & 3.809 / 54.7 \\
40 & 75.70 & 65.60 & 66.10 & 57.2 & 57.0 & 42.6 & 3.975 / 61.4 & 3.873 / 57.8 & 3.909 / 57.8 \\
50 & 71.90 & 65.45 & 70.05 & 52.8 & 60.8 & 56.6 & 4.002 / 61.7 & 3.994 / 61.3 & 4.051 / 62.6 \\
60 & 72.95 & 65.15 & 69.60 & 57.6 & 58.0 & 56.8 & 4.075 / 61.9 & 3.961 / 62.3 & 4.186 / 66.1 \\
70 & 68.50 & 63.95 & 66.45 & 57.4 & 63.6 & 59.6 & 4.092 / 64.0 & 4.078 / 65.4 & 4.137 / 66.6 \\
80 & 68.50 & 63.35 & 69.00 & 58.4 & 62.4 & 57.2 & 4.076 / 62.2 & 4.076 / 64.0 & 4.248 / 68.1 \\
\midrule
Mean & 69.98 & 64.15 & 68.35 & 57.80 & 61.33 & 57.87 & 4.081 / 62.70 & 4.038 / 63.90 & 4.190 / 66.93 \\
\bottomrule
\end{tabular*}
\end{minipage}
\end{strip}

Across Tables~\ref{tab:all-checkpoints} and~\ref{tab:mixed-source-checkpoints}, S/S achieves the highest mean LIBERO-40 and CALVIN scores over updates 60, 70, and 80 among all six joint configurations, including 69.33\% mean CALVIN Len-5 success compared with 66.93\% for CFM, while on MT50 it leads T/T at three of the eight post-update evaluations (including update 80); each mean summarizes three checkpoints from the same run.

\newpage
\paragraph{Evaluation reporting details.}
Evaluation tables and curves are generated from checkpoint rollout records, and training-loss figures from logged optimization metrics and SFT TensorBoard events. Aggregate scores follow Section~\ref{sec:experiments}, and evaluation curves connect measured checkpoints without temporal smoothing.

\clearpage
\subsection{Suite-Level Results and Domain-Loss Weights}
\begin{strip}
\begin{minipage}{\textwidth}
\captionsetup{hypcap=false}
\centering
\includegraphics[width=.86\linewidth]{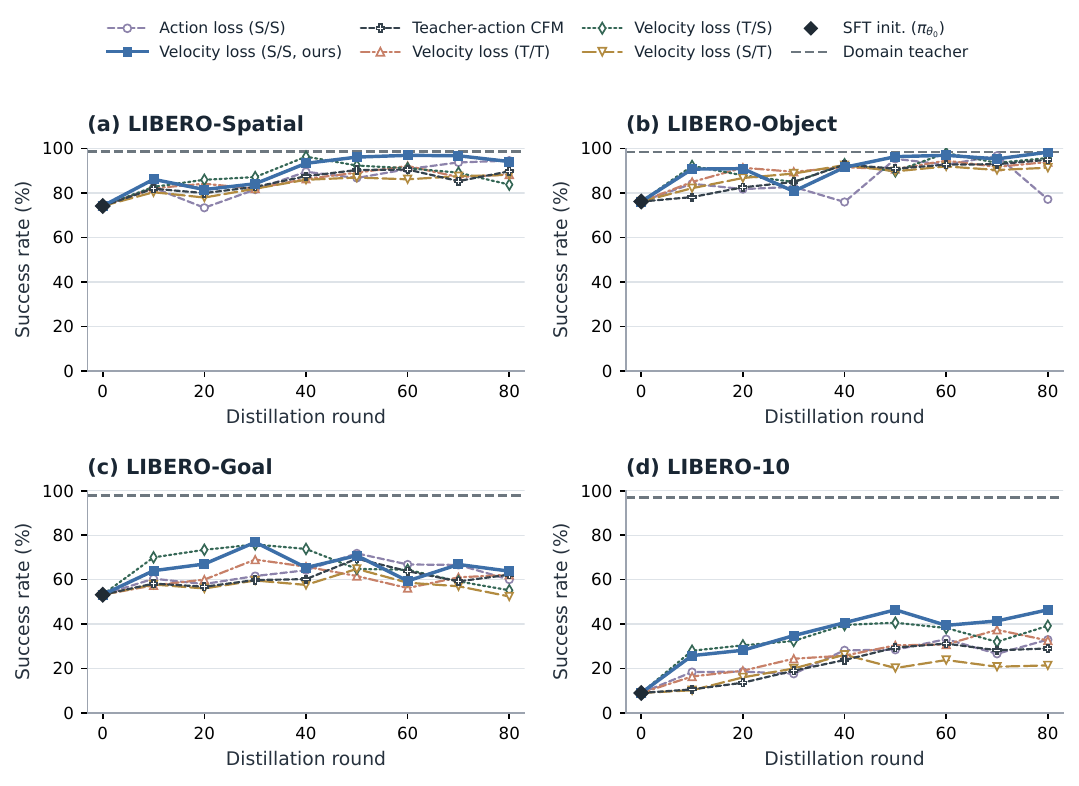}
\captionof{figure}{\textbf{LIBERO success by suite.} Each point averages 500 episodes per suite. Lines follow Figure~\ref{fig:ab-evaluation}. The thick blue curve shows Velocity Loss MOPD (S/S), black diamonds mark the SFT initialization, and gray dashed lines mark the specialists.}
\label{fig:ab-details}
\par\medskip
\captionof{table}{\textbf{Sensitivity to domain-loss weights at round 80.} All three S/S runs share the initialization, optimizer, seed, and 64 episodes per domain per round. Weights are ordered as LIBERO, Meta-World, and CALVIN.}
\label{tab:weight-ablation}
{\small
\setlength{\tabcolsep}{3pt}
\begin{tabular*}{\linewidth}{@{\extracolsep{\fill}}lrrrrrrrr@{}}
\toprule
 & \multicolumn{5}{c}{LIBERO (\%) $\uparrow$} & MT50 & \multicolumn{2}{c}{CALVIN} \\
\cmidrule(lr){2-6}\cmidrule(lr){8-9}
Loss weights & All-40 & Spatial & Object & Goal & LIBERO-10 & SR (\%) $\uparrow$ & Length $\uparrow$ & Len-5 (\%) $\uparrow$ \\
\midrule
$(1/4,1/4,1/2)$ & \textbf{75.70} & 94.2 & \textbf{98.4} & 63.8 & \textbf{46.4} & \textbf{60.6} & \textbf{4.362} & \textbf{72.9} \\
$(1/3,1/3,1/3)$ & 68.05 & \textbf{96.2} & 89.2 & \textbf{69.4} & 17.4 & 57.0 & 4.137 & 65.2 \\
$(1/2,1/4,1/4)$ & 68.85 & 89.2 & 96.0 & 64.0 & 26.2 & 56.4 & 4.024 & 61.8 \\
\bottomrule
\end{tabular*}}

\par\medskip
\captionof{table}{\textbf{Mean scores at updates 60, 70, and 80 for the students in Figure~\ref{fig:single-domain-opd}.} Each single-domain student is evaluated on its own benchmark. Success rates are in percent and CALVIN length is in instructions.}
\label{tab:joint-checkpoint-means}
{\small
\begin{tabular*}{\linewidth}{@{\extracolsep{\fill}}lrrrrr@{}}
\toprule
Configuration & LIBERO-40 & LIBERO-10 & MT50 & CALVIN length & CALVIN Len-5 \\
\midrule
Single-domain (three students) & 71.47 & 31.73 & 53.47 & 4.129 & 64.60 \\
Joint ($1{:}1{:}1$) & 70.98 & 27.40 & 57.87 & 4.114 & 65.17 \\
Joint ($2{:}1{:}1$) & 72.07 & 30.67 & 56.27 & 3.968 & 59.93 \\
Joint ($1{:}1{:}2$) & \textbf{74.67} & \textbf{42.40} & \textbf{57.93} & \textbf{4.238} & \textbf{69.33} \\
\bottomrule
\end{tabular*}}
\end{minipage}
\end{strip}

\clearpage
\subsection{Training Losses}
\begin{strip}
\begin{minipage}{\textwidth}
\captionsetup{hypcap=false}
\centering
\includegraphics[width=\linewidth]{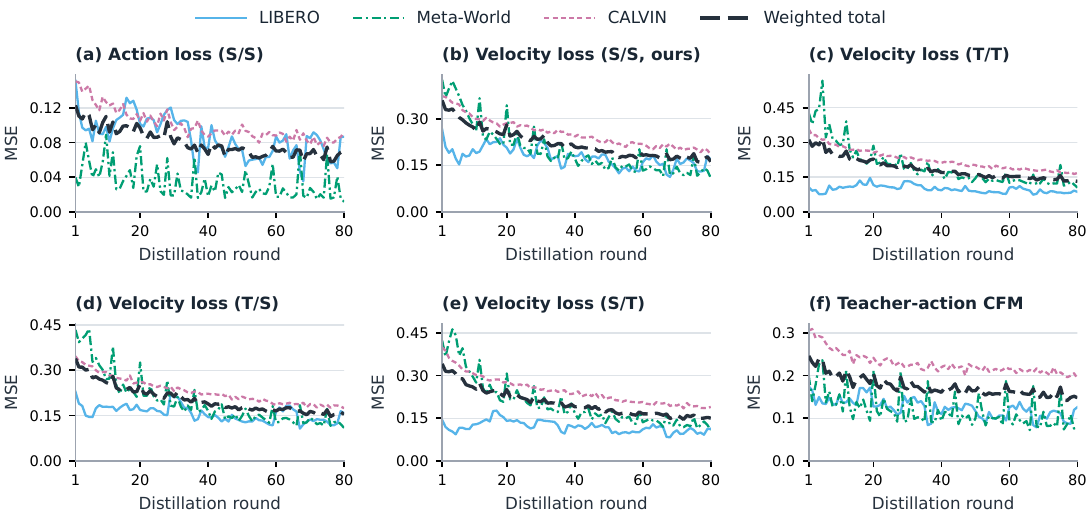}
\captionof{figure}{\textbf{Distillation training losses over 80 rounds.} Unsmoothed domain-average MSE losses for LIBERO, Meta-World, and CALVIN alongside their weighted total (thick long dashes; weights $0.25/0.25/0.50$). Panel (a) matches terminal actions, panels (b)--(e) match teacher velocities across source combinations, and panel (f) uses Teacher-action CFM. Vertical scales reflect the magnitude of each objective and input distribution.}
\label{fig:ab-training}
\end{minipage}
\end{strip}

\begin{strip}
\begin{minipage}{\textwidth}
\captionsetup{hypcap=false}
\centering
\includegraphics[width=.90\linewidth]{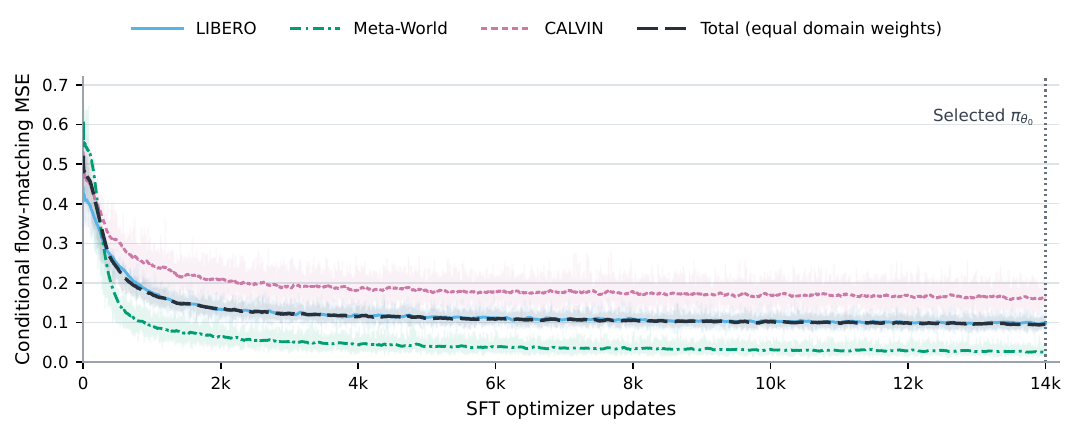}
\captionof{figure}{\textbf{Training the shared SFT initialization.} Conditional flow-matching MSE over the 14,000 optimizer updates that produce $\pi_{\theta_0}$. Solid, dash-dot, and short-dashed curves denote LIBERO, Meta-World, and CALVIN, respectively. Long dashes mark their equally weighted total. Faint traces show raw per-step values, and bold curves show 100-update trailing means.}
\label{fig:sft-training}
\end{minipage}
\end{strip}

\end{document}